\pdfoutput=1
\documentclass[sn-vancouver,Numbered]{sn-jnl}
\usepackage{graphicx}
\usepackage{csquotes}
\usepackage{multirow, makecell}
\usepackage{colortbl}
\usepackage{amsmath,amssymb,amsfonts}%
\usepackage[title]{appendix}%
\usepackage{manyfoot}%
\usepackage{booktabs}%

\usepackage[normalem]{ulem}
\usepackage{dirtytalk}

\begin{document}

\title[Article Title]{GuReT: Distinguishing Guilt and Regret related Text}

\author*[1]{\fnm{Sabur} \sur{Butt}}
\email{saburb@tec.mx}
\equalcont{These authors contributed equally to this work.}

\author[1]{\fnm{Fazlourrahman} \sur{Balouchzahi}}\email{fbalouchzahi2021@cic.ipn.mx}
\equalcont{These authors contributed equally to this work.}

\author[2]{\fnm{Abdul Gafar} \sur{Manuel Meque}}\email{gmequea@cic.ipn.mx}
\equalcont{These authors contributed equally to this work.}

\author[3]{\fnm{Maaz} \sur{Amjad}}\email{maaz.amjad@austin.utexas.edu}
\author[1]{\fnm{Héctor G.} \sur{Ceballos Cancino}}\email{ceballos@tec.mx}
\author[2]{\fnm{Grigori} \sur{Sidorov}}\email{sidorov@cic.ipn.mx}
\author[2]{\fnm{Alexander} \sur{Gelbukh}}\email{gelbukh@cic.ipn.mx}

\affil[1]{\orgdiv{Institute for the Future of Education}, \orgname{Tecnológico de Monterrey}, \orgaddress{\city{Monterrey}, \postcode{64849}, \country{Mexico}}}

\affil[2]{\orgname{Instituto Politécnico Nacional, Centro de Investigación en Computación}, \city{CDMX}, \country{Mexico}}

\affil[3]{\orgname{The University of Texas at Austin}, \city{Austin}, \country{United States}}




\abstract{The intricate relationship between human decision-making and emotions, particularly guilt and regret, has significant implications on behavior and well-being. Yet, these emotions’ subtle distinctions and interplay are often overlooked in computational models. This paper introduces a dataset tailored to dissect the relationship between guilt and regret and their unique textual markers, filling a notable gap in affective computing research. Our approach treats guilt and regret recognition as a binary classification task and employs three machine learning and six transformer-based deep learning techniques to benchmark the newly created dataset. The study further implements innovative reasoning methods like chain-of-thought and tree-of-thought to assess the models’ interpretive logic. The results indicate a clear performance edge for transformer-based models, achieving a 90.4\% macro F1 score compared to the 85.3\% scored by the best machine learning classifier, demonstrating their superior capability in distinguishing complex emotional states.}

\maketitle

\section{Introduction}
Human beings need to make several decisions every day to survive. Some decisions are routine, while others are crucial in determining human behavior. Decisions are essentially a pathway through which emotions shape our everyday efforts to avoid negative sentiments (i.e., guilt and regret) while promoting positive sentiments (i.e., hope, pride, and happiness), even when this process occurs unconsciously~\cite{keltner2010emotion,balouchzahi2023polyhope}. Similarly, decisions can also serve as a channel for intensifying negative emotions or diminishing positive ones linked to mental health issues. Regardless of whether these decisions are advantageous or not, when the consequences of our choices materialize, we typically experience new emotions like elation, surprise, or regret~\cite{lerner2015emotion}. Therefore, emotions and decision-making are intricately interconnected.

Guilt and regret are strongly linked to decision-making because they are emotions that arise in response to our choices and actions. The earliest analysis of regret that took lexicographical, theoretical, and empirical considerations was proposed by Janet Landman~\cite{landman1987regret}, distinguishing it with similar concepts such as shame, guilt, and remorse. 

\begin{displayquote}\textit{Regret is a more or less painful cognitive/affective state of feeling sorry for losses, transgressions, shortcomings, or mistakes. The regretted matters may have been sins of commission as well as sins of omission; they may range from the entirely voluntary to the accidental; they may have been executed deeds or entirely mental ones; they may have been committed by oneself or by another person or group; they may be moral or legal transgressions or morally and legally neutral; and the regretted matters may have occurred in the past, the present, or the future.}
\end{displayquote}

Similarly, an elaborate definition~\cite{landman1987regret} of guilt would have the following: 

\begin{displayquote}\textit{Delinquency or failure in respect
to one’s duty: offense; responsibility for an offense: fault; state of
deserving punishment: deserts; the fact of having committed a breach of
conduct; and the state of consciousness of one who has committed an
offense.}
\end{displayquote}

Based on the definition, we can infer that guilt encompasses two states: (i) states of being, and (ii) states of mind. Moreover, guilt pertains to ethical and legal concerns. The primary challenge is how guilt can be differentiated in natural language. The reason is that these emotions in a limited context might overlap and, in many cases, are indistinguishable. In general, it seems impossible to imagine an instance of guilt without regret; however, it is quite possible to imagine an instance of regret without guilt. Thus, regret is once again the broader concept. Regret cannot be only limited to instances of legal, moral, or psychological culpability, but it also includes instances of legally, morally, and subjectively innocuous acts. Furthermore, unlike guilt, regret cannot be limited to one's own free and voluntary actions and failures to act but also includes the acts and omissions of others and deeds over which one had no control. In contrast to psychology, which employs varying evaluation metrics, natural language lacks explicit indicators of underlying morality and ethics. Consequently, the lexicons for both emotions exhibit significant overlap, and distinguishing between these two emotions within a sentence or paragraph poses a complex challenge. In this paper, we attempt to understand how Natural language processing (NLP) can help us to disambiguate regret and guilt-related text. 



This paper addresses the problem of guilt and regret detection as a binary classification task. Our main contributions are as follows:

\begin{itemize}
    \item We elucidate the textual cues that facilitate the discrimination of guilt from regret (See Section~\ref{Sec: Labelmarkers}).
    \item We created a novel dataset for classifying guilt and regret (See Section~\ref{Section: Dataset}).
    \item We trained and validated the proposed binary classification dataset using machine learning and transformer-based classifiers (See Section~\ref{Section: Methodology}). 
     \item We trained and validated the proposed dataset using three large language models (LLMs) reasoning capabilities: (i) Zero shot, (ii) Few shot Chain-of-thought (CoT), and (iii) Tree-of-thought (ToT)). This is done to assess the reasoning capabilities of large language models in tackling intricate emotion-related tasks. We also present detailed error analyses (See Section~\ref{Section: Error Analyses}). 
\end{itemize}


\section{Regret and Guilt Markers in Text}\label{Sec: Labelmarkers}

\subsection{Outcome of decision}
In accordance with regret theories~\cite{schoeffler1962prediction}, the expected utility associated with making a choice can be expressed as a mathematical function of the probability of the choice multiplied by the value of the choice made, minus the level of regret experienced for not choosing the alternative (the superior option that was not selected). Hence, the outcome of regret always has a degree of uncertainty and it is not easily predictable in both actions and inactions. Whereas guilt~\cite{landman1987regret}, defined as doing something that is below the set standards of the individual, infers that the outcome will be predictable, yet the person still proceeds to do it. This textual marker is visible in many text examples and can be a distinguishing factor. 

\subsection{Focus of emotion}
The focus of the emotional intensity of regret and guilt differ~\cite{mandel2005did}. Although both emotions come from negative decision-making, the focus on emotional negativity that stems from regret emphasizes the negative outcome, whereas guilt emphasizes self-blame and responsibility. 

\subsection{Harm}
It has been proven that guilt and regret both result from the types of harm (i.e., intrapersonal or interpersonal)~\cite{berndsen2004guilt, zeelenberg2008role}. Guilt is primarily associated with scenarios of harm inflicted upon another individual, whereas regret is associated with intrapersonal and interpersonal harm situations. Although when taken alone, harm is a weak indicator of guilt when combined with other textual markers, it becomes a strong indicator of the distinct emotion.

\subsection{Self-discrepancies}
In 1987, Higgins~\cite{higgins1987self} introduced the idea that two separate types of self-discrepancies are connected to distinct emotional experiences. The first type arises from comparing an individual's real self to their ideal self, representing what they aspire to become. This is referred to as an actual/ideal self-discrepancy. The second type of self-discrepancy exists between an individual's real self and what they believe they should be based on societal norms, obligations, and responsibilities, known as actual/ought self-discrepancy.

The research proposed by Davidai and Gilovich~\cite{davidai2018ideal} found that people actually feel more regret about not being the person they could have been than over not being the person they should have been. Regret thus appears to be more related to \textbf{ideal self-discrepancies} than to ought self-discrepancies. While guilt would be associated more closely with \textbf{ought self-discrepancies}~\cite{zhang2021role}. The research also suggested that guilt association with ought self-discrepancies may vary with culture as in a different culture, it also positively correlated with ideal self-discrepancies. Hence, the discrepancies backed by numerous pieces of research can be an ideal text marker for differentiating guilt and regret. 


\subsection{Future decision making}

Regret indicates correctness in future decision-making based on experienced or anticipated regret~\cite{hoerl2016making}. In texts,  modal verbs such as \say{could have,} \say{would have,} and \say{should have} indicate past experiences and hint at altering decisions based on past events. On the other hand, guilt is a painful judgment, and a person committing the act knows the outcome of the decision and knows that it is wrong, yet still decides to do it~\cite{landman1987regret}. Hence, fewer words related to \say{what could have been} are seen. The focus of guilt-related actions then becomes how to alter the situation later, apologize, or escape the reality of it. 

\subsection{Moral Self-blame}
Guilt occurs when an individual concentrates on the particular wrongdoing they committed (\say{I engaged in a wrongful action})~\cite{lickel2014shame} / the focus of the negative energy results in moral self-blaming. This typically occurs when someone has caused harm to a significant relationship and generally instigates efforts to make amends, such as apologizing, rectifying, or reversing the blameworthy behavior. 

This feeling of moral self-blame is significantly higher and more predominant in guilt than regret. The related studies focused on exploring the factors that lead to the emotions of guilt and regret rather than delving into the specific reasons or subjects for which individuals experience regret and guilt. In other words, they looked at what leads people to feel these emotions but did not investigate the specific things that trigger these feelings of regret and guilt.



\section{Literature Review}

The concept of regret, often intertwined with guilt, has been examined through various lenses, shedding light on its multifaceted nature. Regret, in its essence, encompasses cognitive and affective dimensions, with varying degrees of emphasis. It emerges as a result of cognitive appraisals, reflective judgment, and critical thinking, standing as a testament to human rationality. Unlike certain emotions, such as anger or fear, regret leans more towards ``cool" cognitive assessment than ``warm" emotional reactivity. Hence, the anticipation of regret provides a reason to avoid excessive risk-taking \cite{lerner2015emotion}. It frequently accompanies self-reflection, making it an intensely personal experience. Regret's link to the concept of ``possible selves" suggests its potential to invoke emotions and emphasizes its role as an affective phenomenon within the realm of human emotions \cite{landman1987regret}. 

In contrast, guilt, another closely related emotion, often involves a significant degree of arousal and centers around moral concerns and a sense of duty and obligation. Guilt typically arises from a violation of moral standards or a sense of responsibility for an offense \cite{keltner2010emotion}. Regret encompasses a broad spectrum, including both voluntary and involuntary actions, personal and external circumstances, moral and non-moral domains. Unlike regret, guilt is more confined to situations involving transgressions against moral or legal precepts. Solomon's distinction between the two emotions underscores this difference, where guilt entails ``extreme" blame, whereas regret is devoid of blame, as it often concerns circumstances beyond one's control \cite{landman1987regret}. 

Thus, regret and guilt, although sharing certain commonalities in their distressing nature and emotional unpleasantness, possess distinct characteristics and triggers. Regret, characterized by its cognitive and affective dimensions, is a versatile emotion that can encompass a wide range of situations, including those without moral or legal implications. Guilt, on the other hand, is more tightly linked to moral concerns and transgressions against established standards, carrying a heavier sense of duty and responsibility \cite{landman1987regret}. These nuances highlight the complexity of human emotions and showcase how different emotions (e.g., regret and guilt) serve unique roles in shaping human behavior and decision-making \cite{lerner2015emotion}.

\subsection{Techniques and Dataset for Emotions, Regret and Guilt}

Emotion analysis in the realm of digital communication has undergone significant methodological shifts. Early on, emotion analysis was limited in its approach (relay on traditional machine learning algorithms), primarily distinguishing sentiments as positive or negative \cite{li2018text, amjad2022survey}. However, the linguistic diversity and complexity in communications presented new challenges as the digital landscape expanded, and binary classification proved inadequate in capturing the depth of human emotional expression \cite{ahmad2020borrow}. Therefore, more sophisticated models were required for emotion analysis.

Deep learning, as a transformative approach, introduced a multitude of nuanced techniques (e.g., neural networks) for emotion analysis \cite{balouchzahi2023reddit}. A number of neural networks based approaches have been proposed, such as the Deep Averaging Networks (DAN) \cite{iyyer2015deep}, Recurrent Neural Networks (RNN) \cite{giles1994dynamic}, and Convolutional Neural Networks (CNN) \cite{lecun1998gradient}. These foundational models, along with more sophisticated model, such as the Long Short Term Memory networks (LSTM) \cite{hochreiter1997long} and attention mechanisms \cite{itti2001computational}, found applications in diverse platforms. Furthermore, more recent advancements led to the development of transformer-based models (e.g., LLMs), such as Step-by-step reasoning, which further rely on self-consistency of thought~\cite{wang2022self}, chain-of-thought (CoT)~\cite{wei2022chain}, and tree-of-thought (ToT)~\cite{yao2023tree}. Therefore, all these techniques and models (e.g., DialogueGCN \cite{ghosal2019dialoguegcn}, DialogueRNN \cite{majumder2019dialoguernn}, and PAN \cite{rathnayaka2019gated}) exemplify the adaptability, highlight deep learning's efficacy in different communication contexts, and pave the way for a more granular understanding of digital sentiment.

Large Language Models (LLMs) have demonstrated good results in understanding and generating human-like text due to their reasoning abilities \cite{sidorov2023regret}. In the realm of emotion-related tasks, particularly regret and guilt classification, LLMs have been harnessed for their advanced reasoning abilities \cite{butt2021transformer, sidorov2023regret}. There are two methodologies used to enhance the reasoning capabilities of large language models (LLMs): (i) Step-by-step reasoning, and (ii) chain-of-thought (CoT) reasoning. Both methods aim to improve the model's ability to solve complex problems by generating intermediate reasoning steps, but they differ in their approach and implementation\cite{wei2022language}. In step-by-step reasoning, the model breaks down complex emotional nuances into sequential logical steps and generates intermediate reasoning steps automatically \cite{paranjape2023art}. This approach often involves integrating external tools, and it has shown substantial improvements over few-shot prompting and chain-of-thought (CoT) \cite{paranjape2023art, wei2022language, wei2022chain}. This approach allows for a detailed analysis of the emotional context, enabling the LLM to discern subtle distinctions between regret and guilt. 

On the other hand, CoT reasoning is a technique that enables LLMs to articulate a series of intermediate reasoning steps. Unlike Step-by-step reasoning which often integrates external tools and has shown improvements over CoT, CoT prompting guides the model to decompose complex problems into intermediate steps, mimicking an intuitive thought process \cite{wei2022chain}. This method does not require a large training dataset or modifying the language model’s weights, and it is particularly effective when combined with few-shot prompting to capture intricate emotions \cite{wei2022language, wei2022chain}.

Furthermore, the exploration of tree-of-thought (ToT) reasoning in LLMs adds another layer of sophistication to emotion-related tasks. By structuring the reasoning process as a tree, the model can allow exploration over coherent units of text and enable deliberate decision-making through considering multiple reasoning paths and the ability to backtrack when necessary \cite{yu2023towards}. This hierarchical approach aids in capturing the multi-faceted nature of these emotions, allowing for a nuanced and comprehensive classification. Therefore, these methodologies, including CoT, step-by-step reasoning, and ToT, are crucial for understanding and classification of emotions. Table \ref{tab1} summarizes existing emotion analysis models along with their respective datasets. Though a wide variety of emotion-related datasets are available. The majority of them are limited to Ekman~\cite{ekman1999basic} and Plutchik's emotions~\cite{plutchik1984emotions}. Whereas, the current datasets and approaches used to classify regret and guilt exhibit a limited understanding of regret and guilt and overlap several emotions that are distinct in their expression.

\begin{table}[htbp]
    \resizebox{\textwidth}{!}{%
    \begin{tabular}{@{}ccccccccc@{}}
        \toprule
        \multirow{1}{*}{Corpora} & \multicolumn{6}{c}{Ekmans Emotions} & \multirow{1}{*}{Other Emotions} & \multirow{1}{*}{Number} \\ 
        & A   & D   & F   & J   & Sad  & Sur                                                        \\ \hline
        SemEval-2007 & $\checkmark$ & $\checkmark$ & $\checkmark$ & $\checkmark$ & $\checkmark$ & $\checkmark$ & -- & 1250 \\ \hline
        ISEAR & $\checkmark$ & $\checkmark$ & $\checkmark$ & $\checkmark$ & $\checkmark$ & -- & Shame, Guilt & 7600 \\ \hline
        CEASE & $\checkmark$ & -- & $\checkmark$ & $\checkmark$ & $\checkmark$ & -- & \begin{tabular}[c]{@{}l@{}}Guilt, Forgiveness, Pride,\\ Love, Thankfulness, Blame\end{tabular} & 2393 \\ \hline
        NLPCC-2018 & $\checkmark$ & -- & $\checkmark$ & -- & $\checkmark$ & $\checkmark$ & Happiness & 7928 \\ \hline
        EmoInt Dataset & $\checkmark$ & -- & $\checkmark$ & $\checkmark$ & $\checkmark$ & -- & Sentiment with intensity & 7100  \\ \hline
        Ren-CECps & $\checkmark$ & -- & -- & $\checkmark$ & -- & $\checkmark$ & \begin{tabular}[c]{@{}l@{}}Anxiety, Hate, Love, Sorrow,\\ Expect, Sentiment with intensity\end{tabular} & 1487  \\ \hline
        Alm’s Fairy Tale & $\checkmark$ & $\checkmark$ & $\checkmark$ & -- & $\checkmark$ & $\checkmark$ & Happiness, Neural & 1580 \\ \hline
        EEC & $\checkmark$ & $\checkmark$ & $\checkmark$ & $\checkmark$ & $\checkmark$ & $\checkmark$ &\begin{tabular}[c]{@{}l@{}} Love, Optimism, Pessimism, Trust,\\ Anticipation, Sentiment with intensity\end{tabular} & 8640  \\ \hline
        EmotionContext & $\checkmark$ & -- & -- & -- & $\checkmark$ & -- & Happiness, Others & 30k records \\ \hline
        DailyDialog & $\checkmark$ & $\checkmark$ & $\checkmark$ & $\checkmark$ & $\checkmark$ & $\checkmark$ & Neural & 13k dialogues \\ \hline
        MELD & $\checkmark$ & $\checkmark$ & $\checkmark$ & $\checkmark$ & $\checkmark$ & $\checkmark$ & Neural, Positive, Negative & 13k utterances  \\ \hline
        EmoryNLP & -- & -- & -- & $\checkmark$ & $\checkmark$ & -- & Neutral, Mad, Scared, Powerful, Peaceful & 12.6k utterances \\ \hline
        EmotionLines & $\checkmark$ & $\checkmark$ & $\checkmark$ & -- & $\checkmark$ & $\checkmark$ & Neural, Happiness & 2.9k utterances \\ \hline
        SEMAINE Database & -- & -- & -- & -- & -- & -- & Dimensional & 240 dialogues \\ \hline
        IEMOCAP & $\checkmark$ & -- & $\checkmark$ &  & $\checkmark$ & $\checkmark$ & \begin{tabular}[c]{@{}l@{}} Dimensional, Happiness, Excitement, \\Frustration, Other and Neutral\end{tabular} & 5.5k utterances \\ \hline
        EMOBANK & $\checkmark$ & $\checkmark$ & $\checkmark$ & $\checkmark$ & $\checkmark$ & $\checkmark$ & Dimensional&  10k \\ \hline
        ReDDIT & -- & -- &--& -- & -- & -- & \begin{tabular}[c]{@{}l@{}}Regret (by action and inaction),\\ No Regret\end{tabular} &  3425 \\ \hline
        VIC  & -- & -- &--& -- & -- & -- & Guilt, No-Guilt &  4622 \\ 
        \bottomrule
    \end{tabular}%
    }
    \caption{Publicly Available Datasets~\cite{deng2021survey} for Textual Emotion Detection. Ekman’s 6 basic emotions: A-Anger, D-Disgust, F-Fear, J-Joy, Sad-Sadness, Sur-Surprise. All datasets are in English.}
    \label{tab1}
\end{table}

\section{Dataset Development}~\label{Section: Dataset}

In this section, we embark on a comprehensive exploration of the datasets pivotal to our research endeavors. We begin by presenting the Regret Detection and Domain Identification Dataset (ReDDIT) \cite{balouchzahi2023reddit}, followed by an introduction to the Guilt Detection in Text Dataset (VIC)~\cite{meque2023guilt}. After performing thorough pre-processing steps, we tailored the datasets to our specific needs, re-annotated the dataset, and created a unified dataset for guilt and regret detection experiments. We present you the details of annotation and data statistics.

\subsection{Data collection and pre-processing}

The ReDDIT dataset introduced in \cite{balouchzahi2023reddit, sidorov2023regret}, serves as a valuable resource for understanding the expression of regret on social media, particularly on the Reddit platform. This dataset offers a fine-grained analysis of regret, classifying texts into three distinct categories: Regret by Action, Regret by Inaction, and No Regret. The data collection process involved scraping posts from three relevant subreddits, including \say{regret}, \say{regretfulparents}, and \say{confession} spanning from January 1, 2000, to September 10, 2022. The dataset has undergone a meticulous annotation process, with three annotators having backgrounds in information technology and computer science, ensuring its reliability. It provides valuable insights into the domains most commonly associated with regret, especially in the realm of relationships. The dataset's statistics reveal the distribution of posts across different regret types and domains, making it a valuable resource for researchers developing and evaluating natural language processing algorithms focused on detecting and understanding emotional language in online texts.

The VIC dataset, first introduced in~\cite{meque2023guilt} and also featured in~\cite{meque2024leveraging}, is a pioneering resource that addresses the previously understudied area of guilt detection in text using Natural Language Processing (NLP). This dataset, created by combining and binarizing samples from three existing emotion detection datasets, namely VENT~\cite{ventlykousas2019}, ISEAR~\cite{ISEAR-Scherer_1994}, and CEASE~\cite{ceaseghosh2020}, provides a comprehensive foundation for exploring guilt as an emotion in textual content. The dataset development process involved selecting relevant samples from these datasets, emphasizing guilt-labeled categories. Careful data cleaning and balancing techniques resulted in a dataset with a balanced distribution of guilt and no-guilt samples, making it suitable for machine learning experiments. The VIC dataset's statistics showcase its origin and characteristics, highlighting differences in average sentence and word lengths between guilt and no-guilt samples. Researchers can leverage this dataset to delve into the complex realm of guilt detection in the text, laying the groundwork for future research and advancements in understanding this intricate emotion through NLP methods. 

Both dataset were subjected to meticulous preprocessing to address class imbalance and ensure a judicious distribution of samples. For VIC, we initiated the procedure by grouping the dataset by the \say{origin} column, subsequently applying random sampling with replacement to select 500 samples from each group. These operations were followed by a judicious selection of pertinent columns, namely \say{text} and \say{label}, signifying textual content and corresponding labels, respectively. To rectify the class imbalance in ReDDIT and align the dataset with our research objectives, we initiated the process by grouping the dataset by the \say{label}. We then executed random sampling with replacement to select 500 samples from each group. Furthermore, to enhance label interpretability, we employed a label transformation scheme, replacing numeric labels (0, 1, and 2) with descriptive labels, specifically \say{no regret} and \say{regret.}

The culminating step in our dataset preparation involved the concatenation of \say{VIC} and \say{ReDDIT} along the rows, a procedure executed employing. The resultant dataset is distinguished by its balanced composition, affording parity between the various class categories. Table~\ref{tab:Datastats} shows the statistics of the dataset after pre-processing and unification.

\begin{table}[h]
\begin{tabular}{ll}
\hline
Label  & Size \\ \hline
Regret & 755  \\ \hline
Guilt  & 933  \\ \hline
Total  & 1688 \\ \hline
\end{tabular}%
\caption{Data statistics after pre-processing}
\label{tab:Datastats}
\end{table}

\subsection{Annotator selection and procedure}

Three annotators were selected based on their background in multidisciplinary research in psychology and computation, all three held a Masters's degree.  Thirty samples were provided to the qualified annotators for annotation. The annotations were evaluated manually and the common problems in understanding were discussed. One annotator was female, while the other two were male. Each annotator was given a subset of the dataset and was asked to mark the most suitable label according to the provided guideline explained in Section~\ref{Sec: Labelmarkers}.

\subsection{Inter-annotator agreement}
Inter-annotator agreement (IAA) measures the degree of consensus among annotators, taking into account the possibility of random agreement. For the final labels, three annotators achieved a 90.67\% Cohen’s Kappa Score~\cite{banerjee1999beyond}. These scores demonstrate the reliability of the datasets and reflect the rigorous annotation process that was followed. While, for the labels that were previously annotated in the original dataset, we calculated Cohen’s Kappa Coefficient which reached 85.57\% with our final labels after we removed and filtered all instances which were insufficient for the distinction between regret and guilt. 


\section{Methodology}~\label{Section: Methodology}

The section presents the details of deep learning and machine learning models used for binary classification. 
We also employed three prompting techniques in our experimental design using large language models, utilizing GPT-3.5-Turbo\footnote{{\url{https://openai.com/blog/chatgpt/}}}, as our language models (LLM). All LLM experiments were carried out using LangChain~\footnote{{\url{https://python.langchain.com/en/latest/}}}.

\subsection{Pre-Processing}
We followed conventional pre-processing methods, including the removal of duplicate instances, punctuations, URLs, and usernames. Nevertheless, we chose not to incorporate additional pre-processing steps such as stopword removal and lemmatization. This choice was made because language models and transformers rely on sentences in their authentic, unaltered form. 

\subsection{Algorithm Design}
\subsubsection{Machine Learning}
We explored a variety of traditional machine learning algorithms to compare and contrast with the emerging transformer architectures, including large language models. Central to our assessment is the utilization of diverse ensemble learning models which have historically demonstrated robust performance across a plethora of tasks. Specifically, we examine the efficacy of Random Forest \cite{breiman2001random}, AdaBoost\cite{freund1997decision}, and XGBoost\cite{chen2016xgboost} classifiers for the emotion classification task, leveraging their inherent strengths in handling complex decision spaces.

Initial processing involved converting the textual data into a numerical format amenable to machine learning models. We derive vector representations of the text by averaging word embeddings using FastText~\cite{joulin2016fasttext} word vectors, effectively capturing the semantic essence of each sentence within a high-dimensional feature space. This transformation results in embeddings that serve as inputs to our classifiers.

\subsubsection{Transformers}
The advent of Transformer-based models has revolutionized the field of Natural Language Processing (NLP). Leveraging the powerful representational capabilities of these pre-trained models, our study evaluates their performance on the complex task of emotion classification in text. We utilize a suite of Transformer architectures, including BERT \cite{devlin2018bert}, RoBERTa \cite{liu2019roberta}, AlBERT \cite{lan2019albert}, XLNet \cite{yang2019xlnet}, DistilBERT \cite{sanh2019distilbert}, and ELECTRA \cite{clark2020electra}, to probe the breadth and depth of each model's understanding of nuanced emotional constructs. We instantiate a designated Transformer model and fine-tune it to the emotion classification task at hand. The fine-tuning process employs a learning rate of $3 \times 10^{-5}$ over 15 epochs, with sequence lengths and batch sizes carefully calibrated to balance computational efficiency and model performance.

\subsubsection{LLMs}

\begin{enumerate}
    \item Zero-shot prompting: The Zero-Shot \cite{xian2020zeroshot} (ZS) approach aims to directly classify the sentiment of a text as \say{Regret} or \say{Guilt} without explicitly modeling the underlying reasoning process. In this approach, we leverage the powerful pre-trained language model GPT-3.5-Turbo to perform one-shot text classification. We constructed a simple template prompt for GPT-3.5-Turbo:

\texttt{Given a text, identify if it represents \say{Regret} or \say{Guilt}. Only write the final class. \\
Text: \textit{<text\_to\_classify>} \\
Final Class: \textit{output}}\\

We accessed GPT-3.5-Turbo via OpenAI's API, calibrated at temperature 0 for determinism and reduced diversity. The model was not explicitly trained on our data, relying solely on pre-trained knowledge and one-shot generalization.

\item Few-shot Chain of Thought prompting: The Chain of Thought prompting strategy~\cite{wei2022chain}, as applied within our research, constitutes a template-driven approach that impels a language model to simulate a cognitive process akin to human reflection and reasoning. The approach is crafted to distill the model's rationale behind its classification of emotional sentiments into discernible \say{Thoughts,} explicitly delineating a synthesized logical derivation that leads to its final choice between \say{regret} and \say{guilt.} In the prompt format used for CoT, the model is provided with five examples of regret and guilt with their respective thoughts and text instances that capture a human experience or sentiment. Following this, the model is required to classify the sentiment as either \say{regret} or \say{guilt.} Most critically, it must augment its classification with an articulated \say{Thought,} a narrative element that methodically elucidates the reasoning behind its decision. This constructed narrative mirrors the logical steps that individuals typically employ when interpreting complex emotions and providing justifications for their interpretations.


\item Tree of Thought prompting: For this approach~\cite{yao2023tree}, we encoded the key expert knowledge provided in the prompt about differentiating \say{Regret} and \say{Guilt} into seven distinct rules:

 \texttt{\begin{enumerate}
   \item Outcome Source: Regret from the outcome, Guilt from responsibility.
    \item Future Impact: Regret guides future decisions, Guilt is a painful judgment.
    \item Focus: Regret on negative outcome, Guilt on self-blame.
    \item Outcome Predictability: Regret for unpredictable outcomes, Guilt for predictable ones.
    \item Harm Focus: Guilt from interpersonal harm, Regret from both self and others.
    \item Self-Discrepancy: Regret relates to the ideal self, Guilt to the ought self.
    \item Moral Blame: Higher in Guilt than Regret.
\end{enumerate}}

The resulting ToT simulates a round-robin discussion where each expert analyzes the text, applies relevant rules, and explains their reasoning. They build upon each other's contributions, acknowledging and correcting misunderstandings until a consensus on the final class (\say{Regret} or \say{Guilt}) is reached.
\end{enumerate}

\subsection{Evaluation}
Our evaluation strategy for machine and deep learning models employed stratified K-Fold cross-validation with five folds, ensuring that each class's distribution is uniform across each partition. This method lends statistical rigor to our analysis, providing a cross-sectional view of model performance while mitigating sampling variability. Each fold undergoes meticulous processing where the train set is used to fit the model, and the resulting classifier is used to predict labels on the test set. We report on a suite of metrics – accuracy, precision, recall, and F1 score to provide a comprehensive understanding of model behavior. Scores are calculated both with weighted and macro-average options, offering insight not only into the aggregate performance but also into how each model performs in balance and fairness across classes. These metrics collectively give us an opportunity to scrutinize the models' decision-making fabric, benchmarking them against advanced NLP models. For LLMs, all the data was used as the test set. ToT needed manual evaluation to come up with the final answers on many of the instances where the output was not provided as described in the prompted structure

\subsection{Results}
The tables~\ref{tab: ML_results},~\ref{tab: T_results}, and \ref{tab: LLM_results} present the results of the binary classification for machine learning, deep learning, and large language models. We can see that the machine learning models, including Random Forest, AdaBoost, and XGBoost, demonstrate comparable performance with accuracy ranging from 0.840 to 0.845. These models exhibit balanced precision, recall, and F1 scores for both weighted (W) and macro (M) averages. In contrast, the transformer models, such as Bert, Roberta, Albert, Xlnet, Electra, and Distilbert, outperform the traditional machine learning models, achieving accuracy scores between 0.894 and 0.908. These transformers consistently show higher precision, recall, and F1 scores across both weighted and macro averages. Lastly, the language model GPT-3.5-turbo performs variably across different tasks. In zero-shot learning, shows moderate accuracy and F1 scores, while few-shot learning improves accuracy and F1 scores. However, in the task-oriented transfer (ToT) setting, the model's performance drops, indicating an inability to reason for our task. Overall, the transformer models outshine traditional machine learning models and GPT-3.5-turbo in the given evaluation metrics.

\begin{table}
\begin{tabular}{lccccccc}
\hline
Models & P\_W & R\_W & F1\_W & P\_M & R\_M & F1\_M & Acc \\ \hline
Random Forest &	0.843 &	0.840 &	0.839 &	0.845 &	0.833 &	0.836 &	0.840\\ \hline
AdaBoost &	0.842 &	0.841 &	0.841 &	0.840 &	0.840 &	0.840 &	0.841 \\ \hline
XGBoost &	0.846 &	0.845 &	0.845 &	0.845 &	0.843 &	0.843 &	0.845 \\\hline
\end{tabular}
\caption{The table presents the machine learning results. Evaluation metrics are denoted as Acc (Accuracy), Pr (Precision), Rec (Recall), M (Macro), W (Weighted), and F1 (F1 Score).}
\label{tab: ML_results}
\end{table}

\begin{table}

\begin{tabular}{lccccccc}
\hline
Models & P\_W & R\_W & F1\_W & P\_M & R\_M & F1\_M & Acc \\ \hline
Bert & 0.902 & 0.900 & 0.900 & 0.901 & 0.899 & 0.899 & 0.900 \\ \hline
Roberta & 0.895 & 0.894 & 0.894 & 0.895 & 0.892 & 0.893 & 0.894 \\ \hline
Albert & 0.908 & 0.905 & 0.905 & 0.906 & 0.905 & 0.904 & 0.905 \\ \hline
 Xlnet & 0.898 & 0.896 & 0.896 & 0.896 & 0.896 & 0.895 & 0.896 \\ \hline
 Electra & 0.894 & 0.892 & 0.891 & 0.893 & 0.890 & 0.890 & 0.892 \\ \hline
 Distilbert & 0.897 & 0.896 & 0.896 & 0.896 & 0.895 & 0.895 & 0.896 \\ \hline
\end{tabular}
\caption{The table presents the transformers results. Evaluation metrics are denoted as Acc (Accuracy), Pr (Precision), Rec (Recall), M (Macro), W (Weighted), and F1 (F1 Score).}
\label{tab: T_results}
\end{table}

\begin{table}
\begin{tabular}{llccccccc}
\hline
Models&Method & P\_W & R\_W & F1\_W & P\_M & R\_M & F1\_M & Acc \\ \hline
\multirow{3}{*}{Gpt 3.5-turbo}&Zero-Shot&	0.720&	0.549&	0.492&	0.467&	0.392&	0.339&	0.549\\ 
&Few shot CoT&	0.769&	0.616&	0.632&	0.505&	0.429&	0.429&	0.616\\ 
&ToT&	0.601&	0.463&	0.481&	0.398&	0.325&	0.330&	0.463 \\\hline
\end{tabular}
\caption{The table presents the LLM results. Evaluation metrics are denoted as Acc (Accuracy), Pr (Precision), Rec (Recall), M (Macro), W (Weighted), and F1 (F1 Score).}
\label{tab: LLM_results}
\end{table}

\section{Error Analyses and Future Work}\label{Section: Error Analyses}

\begin{figure}[!thbp]
\centering
 \includegraphics[width=.48\linewidth]{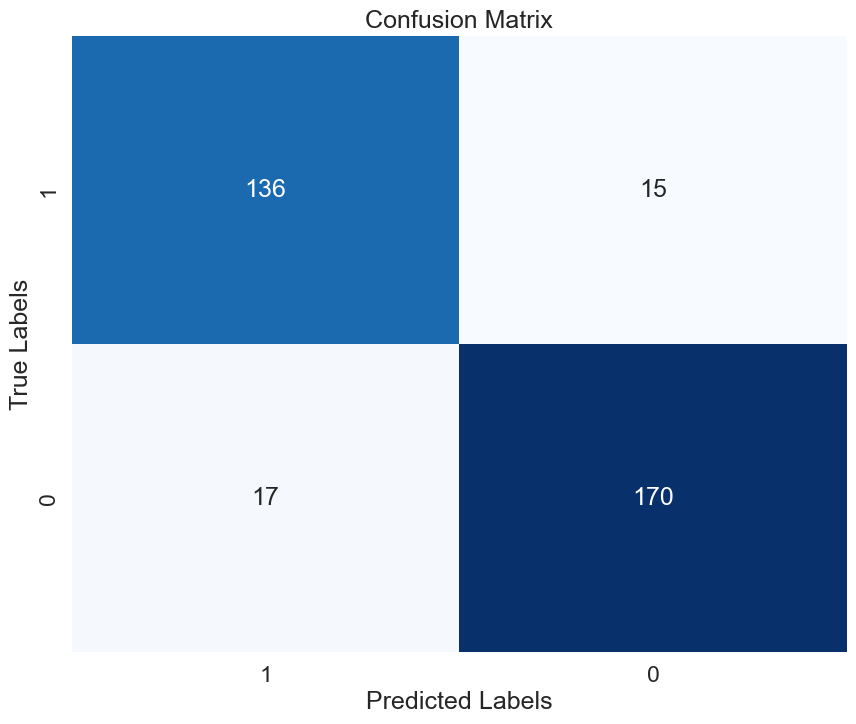} \hfill
 \includegraphics[width=.48\linewidth]{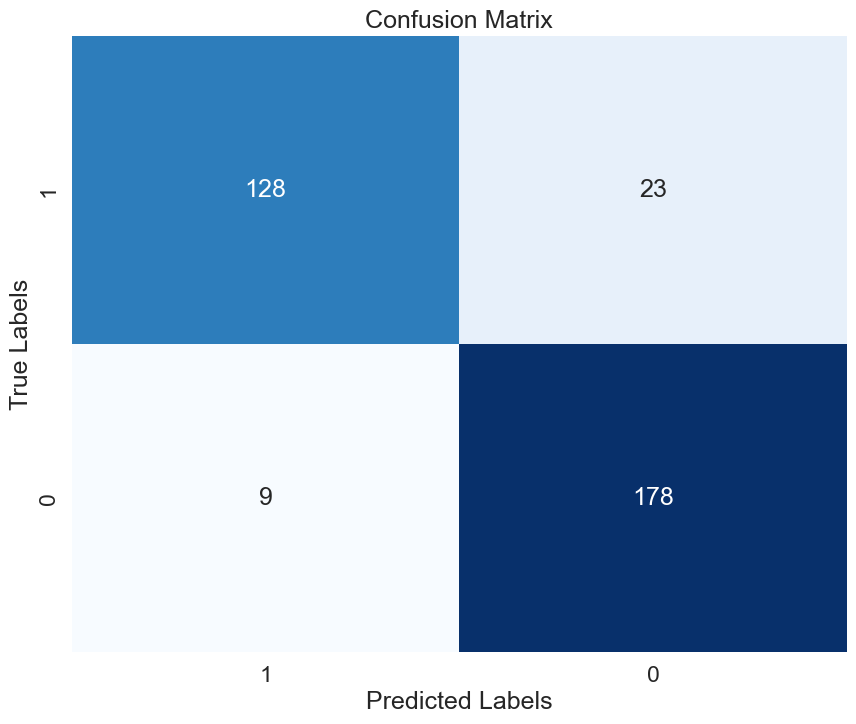} \\
 \vspace{5mm}

 \includegraphics[width=.48\linewidth]{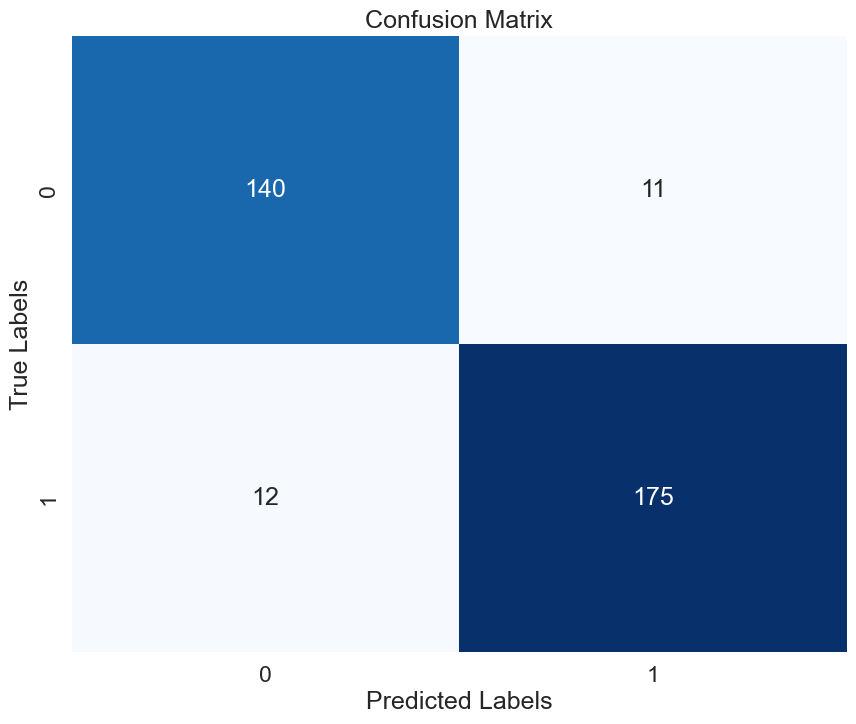}\hfill
 \includegraphics[width=.48\linewidth]{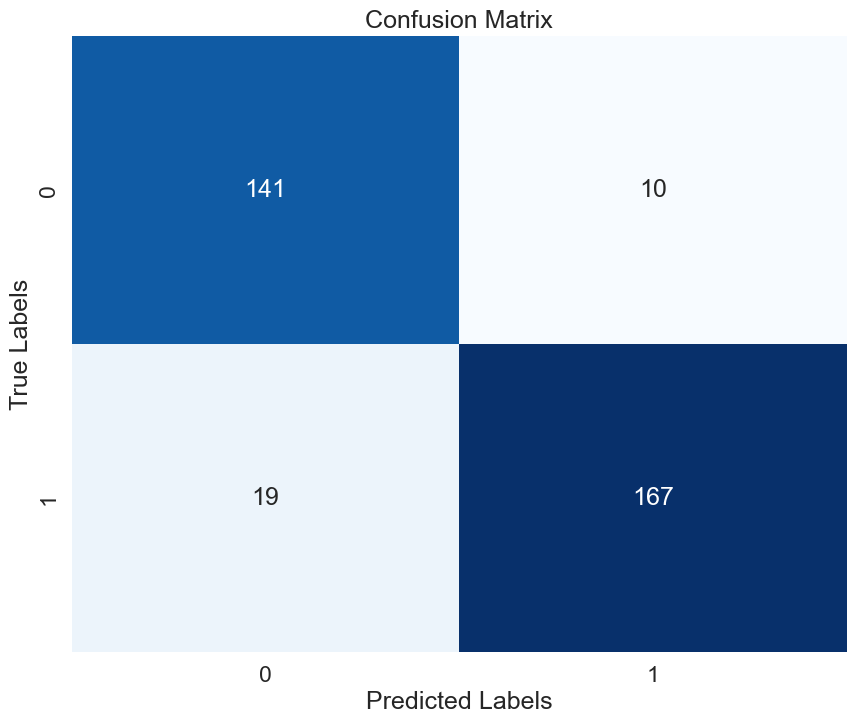}
 \includegraphics[width=.48\linewidth]{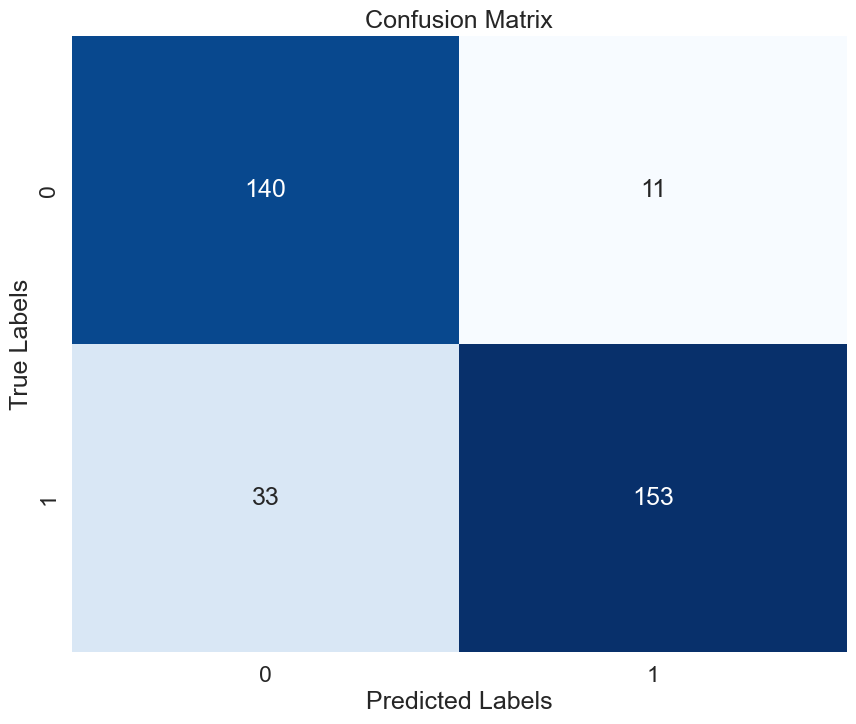}
\caption{Confusion matrix of every 5-fold evaluation with AlBert}
\label{Confusion_matrix_albert}
\end{figure}

We conducted an error analysis on our best-performing model (AlBERT) and large language model (GPT-3.5-Turbo). Figure~\ref{Confusion_matrix_albert} shows the confusion metrics between labels for every fold. In many of the instances of guilt that were identified as regret, the algorithm failed to understand the \say{moral self-blame} indicator. Although compared to Machine Learning, transformers were able to understand the context better to make the distinction. On the other hand, instances of regret that are identified as guilt contain apologetic keywords such as \say{sorry}, however, in reality, do not have strong indicators supporting guilt i.e. \say{I hurt myself again. I stopped using this for a little while, cause of school and stuff, and a crap tom happened while I was gone. I hurt myself, got a crush, confessed to that crush, got good grades at school, and through all of that I was feeling miserable.}, where the focus of emotion is the negative outcome. In contrast, large language models presented several problems in understanding complex contexts. Tree of Thought method, often made a point and then deviated from it, or was inconclusive to which class it should belong to. The ambiguity of the responses also included responses such as \say{i don't know} or \say{difficult to determine}, which in reality are clear examples of their respective classes. The LLM methods also presented repetitive patterns and unstructured responses. All of this made it difficult to evaluate and determine the final class automatically.  
Although we believe, fine-tuning LLM's might counter some of these problems, however, in terms of reasoning abilities LLM's lacked understanding of the complex emotions in our task. 

In the future, we would like to extend the dataset by providing neutral instances and negative emotions that are distinct in their understanding but can overlap with the existing labels. We also want to experiment thoroughly on fine-tuning the existing LLMs to see the potential of LLMs of all sizes and types in understanding these complex emotions.

\section{Conclusion}
Drawing from comprehensive research, empirical analysis, and nuanced discussions on the intricate relationship between emotions and decision-making, this paper underscores the symbiotic connection between emotions and the choices we make. It delves into the depths of human psychology, drawing a fine line between the feelings of guilt and regret that arise from our actions and choices. The paper elucidates the covert indicators that help distinguish between these emotions, which, despite their apparent similarities, emerge from different cognitive processes and moral evaluations.

The paper contributes to the field by successfully mapping textual cues to emotional states, distinguishing between guilt and regret with considerable accuracy. Testing the boundaries of large language models and traditional machine learning algorithms provides compelling evidence for the superiority of transformer-based models in understanding the nuances of emotional language. Furthermore, the development of a novel dataset, tailored annotation processes, and meticulous error analysis add structural integrity to this work.

In closing, this paper not only bridges the gap between theoretical understandings of emotions and their linguistic realizations but also paves the way for technological solutions that can perceive and process complex human emotions with unprecedented depth and sensitivity. The future work proposed promises to extend the reach of this research, inviting more comprehensive explorations and technological advancements that can fine-tune the discernment of AI to interpret emotions as delicately as the human mind does. Within the intricate tapestry of language, emotion, and technology, this paper marks a significant stride in the journey toward a more emotionally intelligent future.

\section*{Declarations}

\begin{description}
    \item\textbf{Funding:} The work was done with partial support from the Mexican Government through the grant A1-S-47854 of CONACYT, Mexico, grants 20241816, 20241819,  and 20240951 of the Secretaría de Investigación y Posgrado of the Instituto Politécnico Nacional, Mexico. The authors thank the CONACYT for the computing resources brought to them through the Plataforma de Aprendizaje Profundo para Tecnologías del Lenguaje of the Laboratorio de Supercómputo of the INAOE, Mexico and acknowledge the support of Microsoft through the Microsoft Latin America PhD Award.

\end{description}

\begin{description}
   \item\textbf{Availability of data and materials:} Data will be made available on request. 
\end{description}









\bibliography{sn-bibliography}

\end{document}